%% file: main.tex
\documentclass[letterpaper, 10pt, conference]{ieeeconf}
\IEEEoverridecommandlockouts
\usepackage{cite}
\usepackage{amsmath,amssymb,amsfonts,mathtools}
\usepackage{algorithmic}
\usepackage{graphicx}
\usepackage{textcomp}
\usepackage{xcolor}
\usepackage{stmaryrd}
\def\BibTeX{{\rm B\kern-.05em{\sc i\kern-.025em b}\kern-.08em
    T\kern-.1667em\lower.7ex\hbox{E}\kern-.125emX}}
\usepackage{algorithm}
\usepackage{url}
\usepackage[nolist,nohyperlinks]{acronym}
\usepackage[hidelinks,bookmarks]{hyperref} 
\usepackage{todonotes}
\usepackage{etoolbox}
\usepackage{siunitx}
\usepackage{booktabs}
\makeatletter
\patchcmd{\@makecaption}
  {\scshape}
  {}
  {}
  {}
\makeatother

\DeclareMathOperator*{\argmin}{arg\,min}

\newcounter{definition}[section]
\newenvironment{definition}[2][]{\refstepcounter{definition}\par\medskip
   \noindent \textbf{Definition~\thedefinition} #1 (#2). \rmfamily}{\medskip}

\newcounter{theorem}[section]
\newenvironment{theorem}[1][]{\refstepcounter{theorem}\par\medskip
   \noindent \textbf{Theorem~\thetheorem} #1. \rmfamily}{\medskip}

\newcommand{\state}[0]{\mathbf{x}}
\newcommand{\ctrl}[0]{\mathbf{u}}
\newcommand{\states}[0]{\mathcal{X}}
\newcommand{\ctrls}[0]{\mathcal{U}}
\newcommand{\safestates}[0]{\mathcal{C}}
\newcommand{\Kappa}[0]{\mathcal{K}}
\newcommand{\Lie}[1]{\mathcal{L}_{#1}}
\newcommand{\pos}[0]{\mathbf{p}}
\newcommand{\vel}[0]{\mathbf{v}}
\newcommand{\acc}[0]{\mathbf{a}}

\newcommand{\local}[1]{\prescript{}{\mathcal{B}}{#1}}

\newacro{fov}[FoV]{Field-of-View}
\newacro{tof}[ToF]{Time-of-Flight}
\newacro{cbf}[CBF]{Control Barrier Function}
\newacro{ecbf}[ECBF]{\textit{Exponential Control Barrier Function}}

\begin{document}

\title{\LARGE \bf  Embedded Safe Reactive Navigation for Multirotors Systems using Control Barrier Functions}

\author{
    Nazar Misyats$^{1,2}$, Marvin Harms$^2$, Morten Nissov$^2$, Martin Jacquet$^2$, Kostas Alexis$^2$
    \thanks{$^1$\'Ecole normale supérieure de Rennes, France, $^2$Autonomous Robots Lab, Norwegian University of Science and Technology (NTNU), Trondheim, Norway,
    {\tt \footnotesize
        \href{mailto:marvin.c.harms@ntnu.no}{marvin.c.harms@ntnu.no}}
    }
    \thanks{This work was supported by the European Commission Horizon Europe grants DIGIFOREST (EC 101070405) and SPEAR (EC 101119774), and by the Research Council of Norway Award NO-321435.}
}

\maketitle

\begin{abstract}

Aiming to promote the wide adoption of safety filters for autonomous aerial robots, this paper presents a safe control architecture designed for seamless integration into widely used open-source autopilots. Departing from methods that require consistent localization and mapping, we formalize the obstacle avoidance problem as a composite control barrier function constructed only from the online onboard range measurements. The proposed framework acts as a safety filter, modifying the acceleration references derived by the nominal position/velocity control loops, and is integrated into the PX4 autopilot stack. Experimental studies using a small multirotor aerial robot demonstrate the effectiveness and performance of the solution within dynamic maneuvering and unknown environments.

\end{abstract}


\input{introduction}

\input{background}

\input{problem_formulation}

\input{method}

\input{implementation}

\input{experiments}

\input{conclusion}

\bibliographystyle{IEEEtran}
\bibliography{refs}

\end{document}

%% file: introduction.tex
\section{Introduction}

Autonomous aerial robots are widely used in both industrial and scientific applications.
As complex missions for monitoring~\cite{acevedo2013cooperative,popovic2020informative}, exploration~\cite{dang2019graph}, and inspection~\cite{zhu2021online} become more common, the need for ensured safety through efficient collision avoidance algorithms becomes ever more important. To that end,  map-based approaches for obstacle avoidance offer an effective solution~\cite{Sucan12, Tranzatto22}, but can be computationally demanding and particularly sensitive to localization errors, map drift, and tracking imperfections, especially in challenging perception conditions~\cite{Ebadi23}.
Another class of methods aims to alleviate some of these problems by removing the need for a consistent global map. These methods incorporate local information into a collision avoidance scheme \cite{Florence18, Gao19, Zhou22}.

On the other hand, control-oriented methods such as safety filters enable provable safety w.r.t. collisions and have therefore been extensively studied \cite{cbf_qp, Wabersich21}.
Such approaches offer computationally efficient last-resort safety mechanisms, and can be employed in combination with other high-level, task-oriented policies, such as volumetric map-based motion planning.
However, the application of these reactive methods using data from exteroceptive sensors remains an open challenge.

In an early work \cite{3d_quadrotor_safe_control}, the authors present a collision avoidance controller for a quadrotor with discrete, moving obstacles. Here, only changes in the thrust command are induced to avoid obstacles, while changes in attitude are enforced to avoid singular configurations.
The authors of \cite{mc_cbf_backstepping} propose a backstepping approach using \acp{cbf} to enforce simple, convex constraints on position, velocity, and rates. Testing was conducted in simulation.
In a recent work \cite{cbf_cone}, the authors develop a new \ac{cbf} construction for obstacle avoidance for moving obstacles using a collision cone approach.
In \cite{softmax_cbf}, the authors propose to combine locally measured obstacle constraints into a single local \ac{cbf}, and demonstrate the applicability of the proposed approach on a simulated quadrotor in a 2D environment.
The work in \cite{learning_cbf} proposes a full-stack perception-driven feedback controller using two jointly-trained neural networks that estimate a \ac{cbf} and a Control Lyapunov Function (CLF) directly on the observation space of a robot.
A similar approach was proposed by \cite{neural_cbf} and experimentally tested for indoor and outdoor flights with a quadrotor.
In a recent work \cite{dyn_cbf_mpc}, the obstacles are modeled as a set of ellipsoids whose trajectory is estimated over time from LiDAR observations. Finally, the local trajectory planner leverages a dynamic \ac{cbf}-based MPC to handle collision avoidance.

Aiming to facilitate the adoption of such low-level safety filters, this paper contributes a safe control architecture designed to be embedded on commonly used flight controller boards and specifically PX4\footnote{https://px4.io}.
Building upon our previous work in~\cite{composite_cbf_nav}, in which we introduced a computationally scalable safety filter, the presented contribution formalizes the avoidance problem as a \ac{cbf}, computed through onboard range measurements.
Furthermore, the proposed \ac{cbf} design is adapted to the standard cascaded control architecture found in PX4 and other open-source autopilots\footnote{https://ardupilot.org}, and ensures collision avoidance by altering the acceleration references derived by the nominal position/velocity control loops.
The embedded safety filter is evaluated in a set of two experiments with a small-scale quadrotor, demonstrating successful collision avoidance during trajectory tracking as well as against adversarial commands.
A proposed implementation is open-sourced in a public fork of the PX4 stack, alongside its realization in the ModalAI electronics to be found at {\tt \url{https://github.com/ntnu-arl/PX4-CBF}}.

The remainder of the paper is structured as follows. Section~\ref{sec:bg} provides an overview on \ac{cbf} theory, Section~\ref{sec:problem} formalizes the problem formulation, while Section~\ref{sec:method} presents the proposed methodology. Section~\ref{sec:impl} details the embedded implementation of the safety filter, and Section~\ref{sec:results} presents experimental validation, before concluding in Section~\ref{sec:conclusions}.

%% file: background.tex
\section{Preliminaries}\label{sec:bg}

\subsection{Notation}
\begin{tabbing}
 \hspace*{2.2cm} \= \kill
  $\mathbf{x} \in \mathcal{X},  \mathbf{u} \in \mathcal{U}$ \>  state \& control input vectors \\[0.5ex]
  $\mathbf{p},  \mathbf{v}, \mathbf{a} \in \mathbb{R}^3$ \>  position, velocity and acceleration vectors \\[0.5ex]
  $\mathcal{L}_f h, \mathcal{L}_g h \mathbf{u}$ \>  Lie derivatives of $h$ along $f$, $g \mathbf{u}$ \\[0.5ex] 
  $\mathcal{I}$, $\mathcal{B}, \mathcal{V}$ \>  inertial, body and vehicle frames \\[0.5ex] 
  $\mathbf{I}_n$, $\mathbf{0}_{n}$ \>  $n$ by $n$ identity and zero matrix \\[0.5ex] 
  $\dot{h}$ \>  time derivative of $h$ \\[0.5ex] 
  $s'$ \>  derivative of $s : \mathbb{R} \rightarrow \mathbb{R}$ \\[0.5ex] 
  $\|\cdot\|$ \>  the Euclidean norm \\[0.5ex] 
  $\prescript{}{\mathcal{A}}{\mathbf{b}}$ \>  vector $\mathbf{b}$ expressed in frame $\mathcal{A}$ \\[0.5ex]
  $\mathbf{R}_{\mathcal{A}\mathcal{B}}$ \>  rotation matrix from frame $\mathcal{B}$ to frame $\mathcal{A}$ \\[0.5ex]
\end{tabbing}
When not specified, physical vectors are implicitly expressed in the inertial frame, i.e. $\mathbf{b} = \prescript{}{\mathcal{I}}{\mathbf{b}}$.

\subsection{Control Barrier Functions}

We consider any control-affine system
described by
\begin{equation}
\label{eqn:dynamics}
\dot{\state} = f(\state) + g(\state)\ctrl
\end{equation}
with $\state \in \states \subseteq \mathbb{R}^n$ the time-dependent system's state,
$\ctrl \in \ctrls \subseteq \mathbb{R}^m$ the time-dependent input,
$f : \mathbb{R}^n \rightarrow \mathbb{R}^n$ and
${g : \mathbb{R}^n} \rightarrow \mathbb{R}^{n\times m}$
two continuous functions. A variety of robot configurations can be described in the form of (\ref{eqn:dynamics}).

\begin{definition}{Control Invariant Set}
A set $\safestates \subseteq \states$ is said to be
\textit{forward control invariant} for the system (\ref{eqn:dynamics})
if for any initial state $\state_0 \in \safestates$,
there exists a control input trajectory $\ctrl(t) \in \ctrls$
such that the solution $\state(t)$ to (\ref{eqn:dynamics})
with initial state $\state(t_0) = \state_0$ under the input trajectory $\ctrl(t)$
remains in $\safestates$ indefinitely, i.e. $\forall t \ge t_0, \state(t) \in \safestates$.
\end{definition}

\begin{definition}{Control Barrier Function~\cite{cbf_theory_app}}
Consider the function $h : \states \to \mathbb{R}$ such that
$\safestates = \{\state \in \states \ | \ h(\state) \ge 0 \}$ with the property $\| \Lie{g}h \| \neq 0$ whenever $h=0$.
Then $h$ is said to be a \ac{cbf}
if there exists an extended class $\Kappa$ function
$\alpha$ 
such that
\begin{equation}
\label{eqn:cbf_condition}
    \sup_{\ctrl \in \ctrls} \left[
    \underbrace{\Lie{f}h(\state) + \Lie{g}h(\state)\ctrl}_{\dot{h}(\state)}
    \right]
    \ge - \alpha(h(\state)).
\end{equation}
\end{definition}

%

\begin{definition}{Safety}
A state $\state \in \states$ is said to be \textit{safe} whenever $\state \in \safestates$. A control input $\ctrl \in \ctrls$ is said to be safe for a given state $\state \in \safestates$ and \ac{cbf} $h$, when it satisfies equation \eqref{eqn:cbf_condition}.
\end{definition}

From the above definitions, it becomes clear that the existence of a \ac{cbf} $h$ provides a formal certificate on the control invariance of the safe set $\safestates$. Apart from classifying pairs of states and inputs as safe or unsafe, \acp{cbf} are utilized to synthesize a safe control input $\ctrl^*$ from any nominal, unsafe control law $\mathbf{k}(\state)$ in a minimally-invasive manner. This can be achieved using reactive safety filters, originally proposed in \cite{cbf_qp}. The control affine system in \eqref{eqn:dynamics} together with the linear constraint \eqref{eqn:cbf_condition} result in a quadratic program to solve the minimally-invasive safe control problem with a \ac{cbf} $h$

\begin{equation}\label{eqn:filtering}
\begin{split}
    \ctrl^* = & \argmin_{\ctrl \in \ctrls}
        \| \ctrl - \mathbf{k}(\state) \|_2^2 \\
    & \textrm{s.t.} \ \Lie{g}h(\state)\ctrl
        \ge - \Lie{f}h(\state) - \alpha(h(\state)) \text{.}
\end{split}
\end{equation}
However, condition (\ref{eqn:cbf_condition}) --- and the filtering
procedure (\ref{eqn:filtering}) ---
rely on the fact that $h$ is of relative degree 1, i.e.
$\Lie{g}h(\state) \neq \mathbf{0}$. Multiple approaches
to handling \acp{cbf} with higher relative degrees have been proposed. Subsequently, an executive summary of exponential control barrier functions is provided.

\subsection{Exponential control barrier functions}

Let $h : \states \to \mathbb{R}$ be of relative degree $r \ge 1$, i.e.
$\Lie{g}\Lie{f}^{k}h(\state) = \mathbf{0}$ for $k = 0, ..., r-2$
and $\Lie{g}\Lie{f}^{r-1}h(\state) \neq \mathbf{0}$.
Given $r$ negative constants (poles) $p_0, ..., p_{r-1}$,
we define a series of functions $\nu_0, ..., \nu_{r}$ as
$$
\begin{aligned}
     \nu_0 &= h \\
     \nu_1 &= \dot{\nu}_0 - p_0 \nu_0 \\
     &~\vdots \\
     \nu_r &= \dot{\nu}_{r-1} - p_{r-1} \nu_{r-1}
\end{aligned}
$$
and their respective safe sets
as $\safestates_i = \{ \state\in\states \ | \ \nu_i(\state) \ge 0 \}$.
If $\nu_r$ is a \ac{cbf}, 
then $h$ is said to be an \ac{ecbf}.

\begin{theorem}
    If $\safestates_r$ is forward-invariant, then $\safestates_0$
    is forward-invariant for any initial state $\state_0 \in \bigcap_{i=0}^r \safestates_i$~\cite{exp_cbf}.
\end{theorem}

\acp{ecbf} provide a systematic way of constructing
a \ac{cbf} of relative degree 1 from a simple constraint function. This allows us to apply the condition \eqref{eqn:cbf_condition} to the function $\nu_{r-1}$. However, the above theorem poses requirements on the initial conditions, where we assume these are satisfied for the remainder of this work. We refer to \cite{exp_cbf} for further details.

\subsection{Composite control barrier functions}

\begin{figure}
    \centering
    \includegraphics[width=0.98\linewidth]{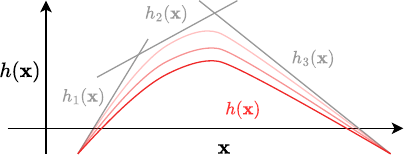}
    \caption{Graphical explanation of the composite CBF $h(\state)$ constructed from $h_i(\state)$ for $i \in \{1,3\}$. $h(\state)$ is a smooth under-approximation of the function $\min_i \hspace{1pt} h_i(\state)$. For increasing the value of $\kappa$ (shown with decreasing opacity of $h$), the approximation becomes less conservative.}
    \label{fig:composite_cbf}
\end{figure}

Enforcing multiple safety constraints is challenging, especially for small aerial robots with limited computational resources. Therefore, in this work we focus on the composition
of multiple \acp{cbf} via a \textit{soft mininimum} function, as proposed in \cite{composing_cbfs}.
Given $n$ \acp{cbf} $h_1, ..., h_n$,
we want to satisfy the invariance condition \eqref{eqn:cbf_condition}  for all of them jointly. We combine each \ac{cbf}
into a single function
$h$, parametrized by $\kappa > 0$, and defined as
\begin{equation}\label{eqn:composite_cbf}
h(\state) \triangleq
-\frac{1}{\kappa}\ln \sum_{i=1}^n \exp\left(-\kappa h_i(\state)\right).
\end{equation}
$h$ is then a CBF if $\kappa$ is chosen large enough. See \cite{composing_cbfs} for a more detailed discussion.
A graphical depiction of the composite \ac{cbf} is shown in Fig. \ref{fig:composite_cbf}.


%% file: problem_formulation.tex
\section{Problem Formulation}\label{sec:problem}

\begin{figure}
    \centering
    \includegraphics[width=0.98\linewidth]{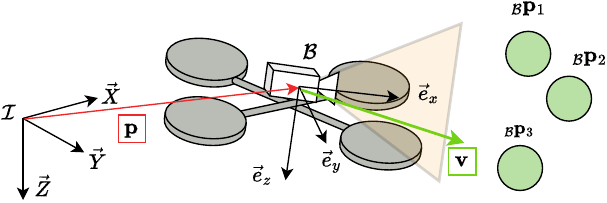}
    \caption{Problem definition and frame conventions used in this work. The multirotor aerial robot seeks to avoid all visible obstacles present in the current sensor measurement. The inertial frame is denoted as $\mathcal{I}$, while the body frame is denoted as $\mathcal{B}$.}
    \label{fig:quad_conventions}
\end{figure}

In this work, we seek to enable collision-free navigation of multirotor aerial robots in unknown environments. In particular, we do not assume any knowledge of a consistent global map nor online reconstruction of such a map from sequential observations.
Instead, the method relies entirely on instantaneous measurements of an on-board range sensor (e.g., stereo camera, \ac{tof} or LiDAR). While multirotors are commonly modeled as a cascaded system of degree 4, the nonlinear attitude dynamics can be controlled by an inner-loop controller. We therefore approximate the translation dynamics as a linear system of the form

\begin{equation}    
\begin{bmatrix}
    \dot{\pos} \\
    \dot{\vel}
\end{bmatrix}
= \underbrace{\begin{bmatrix}
    \mathbf{0}_{3} & \mathbf{I}_{3} \\
    \mathbf{0}_{3} & \mathbf{0}_{3}
\end{bmatrix}
\overbrace{\begin{bmatrix}
    \pos \\
    \vel
\end{bmatrix}}^{\state}}_{f(\state)}
+
\underbrace{\begin{bmatrix}
    \mathbf{0}_{3} \\
    \mathbf{I}_{3}
\end{bmatrix}}_{g(\state)}
\acc,
\end{equation}
where we directly control the acceleration $\acc = \left[a_x \ a_y \ a_z \right]^\top$. The aerial robot operating in an unknown environment receives periodic point measurements of the seen obstacles in the environment from an onboard ranging sensor. The density of measurements and the \ac{fov} of the sensor are assumed to be sufficient for capturing the relevant geometry of the environment in terms of obstacle avoidance, despite angular displacements of the sensor due to the attitude dynamics. At a given time, the sensor measures $n$ obstacles, given by their position in the body frame $\prescript{}{\mathcal{B}}{\pos_1}, ..., \prescript{}{\mathcal{B}}{\pos_n} \in \mathbb{R}^3$.
We want to ensure that the drone does not get closer than $\varepsilon > 0$
from each obstacle, i.e. $\|\prescript{}{\mathcal{B}}{\pos_i}\| \geq \varepsilon \hspace{5pt} \forall i \in \{1\dots n\}$. An illustration of the problem setting and frame conventions is shown in Fig.~\ref{fig:quad_conventions}.

\begin{figure}
    \centering
    \includegraphics[width=0.5\columnwidth]{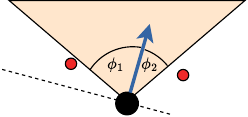}
    \caption{Simplified planar navigation example with a constrained field of view. It illustrates that the vehicle (black) with instantaneous velocity $\vel$ will approach the obstacles (red) located just outside the \ac{fov} if the angles $\phi_1$, $\phi_2$ are less than $\pi/2$. }
    \label{fig:fov_proof}
\end{figure}

It should be emphasized that completely safe, mapless navigation with a sensor with a \ac{fov} under 180$^\circ$ cannot be achieved considering arbitrary motions. This is due to the possibility of unseen obstacles violating approaching the vehicle and violating the constraints. An intuitive explanation for a simplified 2D case are shown in Fig. \ref{fig:fov_proof}.

To still enable collision avoidance with the available information with constrained \ac{fov} sensors, we additionally require the current velocity to be contained within the sensor frustrum. While this choice is conservative, it reduces the risk of colliding with unseen obstacles.

%% file: method.tex
\section{Mapless collision avoidance for UAVs}\label{sec:method}

In this section, we detail the development of a safety filter for multirotor aerial robots. The safety filter acts on the acceleration setpoint
within a standard cascaded position-attitude controller as in widely-utilized open-source autopilots~\cite{meier2015px4} and established literature~\cite{mahony2012multirotor}. In such an architecture, the position/velocity controller typically provides acceleration and yaw commands which are then translated to attitude and thrust references. The attitude references are tracked by the attitude/angular rate controller, and its outputs --combined with the collective thrust-- are sent to the mixer for allocation to the vehicle's motors. 
In this section, we design an optimization
procedure that filters a (potentially unsafe) nominal acceleration setpoint $\acc_\textrm{sp}$ into a new (safe) acceleration setpoint $\acc^*$.

\subsection{Obstacle avoidance constraint}

We utilize a purely reactive approach that uses the real-time
local point cloud from a low-resolution range sensor to avoid
the obstacles directly visible.
At a given time, the robot receives from such a low-resolution sensor a measurement of $n$ range observations which are treated as the centroids of spherical obstacles expressed in $\mathcal{B}$, with corresponding
coordinates $\pos_1, ..., \pos_n$ in the inertial frame.

We make use of a composite \ac{cbf} for expressing the $n$ corresponding constraints into a single, lumped constraint.
In particular, we adapt the formulation introduced in~\cite{composite_cbf_nav} to acceleration control.

We define the $n$ position constraints $\nu_{1, 0}\dots, \nu_{n, 0}$ as

\begin{equation}\label{eqn:cbf_nu0}
\nu_{i, 0}(\state) \triangleq \|\pos_i - \pos \|^2 - \varepsilon^2,
\end{equation}
each $\nu_{i, 0}$ being of relative degree 2.
We introduce the \acp{ecbf}

\small
\begin{equation}
\begin{aligned}\label{eqn:cbf_nu1}
    \nu_{i, 1}(\state) & \triangleq \dot{\nu}_{i, 0}(\state) - p_0 \nu_{i, 0}(\state)
    = -2\vel^\top (\pos_i - \pos)
        - p_0 \nu_{i, 0}(\state).
\end{aligned}
\end{equation}
\normalsize
We then adapt (\ref{eqn:composite_cbf}) to define a CBF over $n$ obstacles as

\begin{equation}\label{eqn:cbf_def_with_sat}
h(\state) \triangleq
    -\frac{\gamma}{\kappa} \ln
    \sum_{i=1}^n
    \exp\left[-\kappa s\left(\nu_{i,1}(\state)/\gamma\right)\right]
\end{equation}
where $s : \mathbb{R} \to \mathbb{R}$ is a saturation function,
here chosen as ${s = \tanh}$ as proposed in \cite{composing_cbfs},
and whose sensitivity is controlled by $\gamma \ge 1$.

The corresponding Lie derivatives
are therefore

\begin{align}
    \Lie{f}h(\state) &=
        \frac{1}{\Lambda(\state)} \sum_{i=1}^n
        \lambda_i(\state) \Lie{f}\nu_{i,1}(\state), \\
    \Lie{g}h(\state) &=
        \frac{1}{\Lambda(\state)}
        \sum_{i=1}^n \lambda_i(\state) \Lie{g}\nu_{i,1}(\state).
\end{align}
where $\lambda_i(\state) = \exp\left[-\kappa s\left(\nu_{i,1}(\state)/\gamma\right)\right] s'(\nu_{i,1}(\state)/\gamma)$,
and $\Lambda(\state)$ is the inner sum of the logarithm in (\ref{eqn:cbf_def_with_sat}).

Furthermore, we have
\begin{align}\label{eqn:cbf_lie_nu1}
    \Lie{f}\nu_{i,1}(\state) &= 
        2\vel^\top\left( \vel + p_0(\pos_i - \pos)\right), \\
    \Lie{g}\nu_{i,1}(\state) &=
        -2 (\pos_i - \pos)^\top.
\end{align}

To enforce the invariance condition (\ref{eqn:cbf_condition}), we choose the extended class $\mathcal{K}$ function in (\ref{eqn:cbf_condition})
 as the function 
 \small
\begin{equation}
\alpha(h) = 
\begin{cases}
    \alpha h & \text{if } h \geq 0\\
    \frac{h}{1/\alpha + |h|}  & \text{otherwise.}
\end{cases}
\end{equation}
\normalsize

\subsection{Field-of-View constraint}

To enforce the \ac{fov} constraints, we choose the constraint functions
\begin{equation}
    c_j (\state) = \prescript{}{\mathcal{B}}{\mathbf{e}}_j^\top \mathbf{R}_{\mathcal{B}\mathcal{I}} \,{\vel} \quad \forall j \in \{1\dots4\},
\end{equation}
where $\mathbf{e}_j$ is a unit vector representing the inward-facing normal on the plane bounding the camera frustum and $\mathbf{R}_{\mathcal{B}\mathcal{I}}$ is the rotation matrix from $\mathcal{I}$ to $\mathcal{B}$. However, this constraint is of mixed relative degree for underactuated aerial robots, such as flat multirotors, since the acceleration is directly controlled by adjusting thrust and $\mathbf{R}_{\mathcal{B}\mathcal{I}}$.
This can also lead to feasibility and stability issues during deployment. We instead use an approximation of the above constraint, where we use the yaw-aligned vehicle frame $\mathcal{V}$ to address these challenges.
This approximation holds for small roll and pitch angles
encountered in non-agile maneuvers, and remains as

\begin{equation}
    h_{fj} (\state) = \prescript{}{\mathcal{V}}{\mathbf{e}}_j^\top \mathbf{R}_{\mathcal{V}\mathcal{I}} \,{\vel} \quad \forall j \in \{1\dots4\}.
\end{equation}
Taking the time derivative, we obtain

\begin{equation}\label{eqn:fov_derivative}
        \dot{h}_{fj} (\state)
        =
        \prescript{}{\mathcal{V}}{\mathbf{e}}_j^\top \mathbf{R}_{\mathcal{V}\mathcal{I}} \,{\acc} + 
        \prescript{}{\mathcal{V}}{\mathbf{e}}_j^\top \dot{\mathbf{R}}_{\mathcal{V}\mathcal{I}} \,{\vel} 
\end{equation}
which is already in the form suited for \eqref{eqn:cbf_condition} with ${\Lie{g}h_j = \prescript{}{\mathcal{V}}{\mathbf{e}}_j^\top \mathbf{R}_{\mathcal{V}\mathcal{I}}}$.
Although the second term on the right-hand side of \eqref{eqn:fov_derivative} 
are in general nonzero, the term remains small in magnitude due to the small yaw rates encountered in multirotor UAVs. We therefore neglect its contribution and assume $\Lie{f}h_j=0$. We note that while the vertical constraints on the field of view can be implemented with this approach, the imposed constraints on the vertical motion can become prohibitive for many tasks. We therefore only consider two horizontal constraints in the remainder of this work.

\subsection{Fully constrained problem}

To enforce all constraints jointly, we implement the minimally-invasive control problem as a soft-constrained QP, where the \ac{fov} constraints are implemented as soft constraints. This avoids feasibility issues in the QP due to imperfect acceleration tracking, noisy observations, and state estimates. The resulting optimization problem, omitting the reference frame $\mathcal{B}$ for legibility, takes the form

\small
\begin{subequations}\label{eqn:cbf_qp_with_fov}
\begin{align}
    \acc^* =&  \argmin_{\acc \in \mathbb{R}^3}  \quad (\acc - \acc_{\textrm{sp}})^\top \mathbf{H} (\acc - \acc_{\textrm{sp}}) + \sum_{i=1}^2 \rho \delta_{fi}  \\ 
     \textrm{s.t.} \hspace{2pt} \Lie{g}h(\state)\acc \ge & - \Lie{f}h(\state) - \alpha(h(\state)) \\
        \Lie{g}h_{fi}(\state)\acc \ge & - \Lie{f}h_{fi}(\state) - \alpha_f h_{fi}(\state) - \delta_{fi} \hspace{3pt} \forall i \in  \{1,2\}\\
        \delta_{fi} \ge & \ 0  \hspace{114pt}   \forall i \in  \{1,2\}
\end{align}
\end{subequations}
\normalsize
for the positive-definite matrix $\mathbf{H}$ and slack multiplier $\rho > 0$.

%% file: implementation.tex
\section{Embedded Implementation}\label{sec:impl}

\subsection{Safe controller architecture}

\begin{figure}
  \centering
  \includegraphics[width=0.5\textwidth]{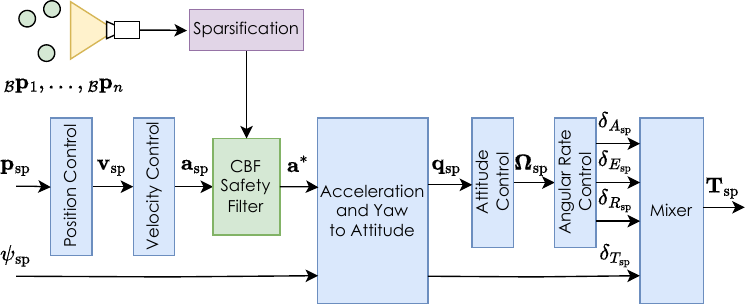}
  \caption{Cascade control scheme diagram with the safety filter introduced.}
  \label{fig:architecture}
\end{figure}

The safety filter is inserted as an intermediate step in the standard cascaded control architecture
of the PX4 autopilot. The filter receives the nominal acceleration setpoint $\acc_\textrm{sp}$
from the velocity controller and a set of $n$ measured obstacle positions
$\local{\pos}_1, ..., \local{\pos}_n$. Using the current velocity and obstacle positions, the composite CBF \eqref{eqn:cbf_def_with_sat} and the corresponding Lie derivatives are computed. A safe acceleration setpoint $\acc^*$ is then computed by solving
the QP in (\ref{eqn:cbf_qp_with_fov}) inside a safety filter before forwarding it to the attitude controller. The full control scheme
is depicted in Fig. \ref{fig:architecture}. 
In the following paragraphs, we detail the implementation of the core
elements of the safety filter, and provide insights on parameter tuning.

\subsection{Obstacles measurement}

Obstacles are measured as a point cloud expressed in $\mathcal{B}$ from the onboard range sensor.
To maintain efficient performance on embedded autopilots and adhere to low-bandwidth constraints, the initial point cloud is sparsified (down to, e.g., $100$ points) before being published to a dedicated PX4 uORB topic using a custom message format.
This down-sampling process is performed by computing the minimum range within angular bins, and selecting the closest points such that the number of obstacles remains within the safety filter’s predefined limit.

The measurements are assumed to arrive at a sufficiently high frequency to prevent significant noise in the \ac{cbf} constraint. Our evaluation indicates that a rate of $10$ Hz or higher is adequate.
Further, the number of points is assumed to be sufficiently high to represent the surrounding environment. We found 100 points to be sufficiently expressive with varying obstacles but fewer points are possible.
Due to framework limitations, each message can store a maximum of $20$ points, with coordinates represented as $32$-bit floating-point values. As a result, the sparsified point clouds are divided into multiple chunks, transmitted via uORB messages, and reconstructed in the receiving module.
On the receiving end, only one of the queued message is read at each iteration to prevent locking.
The points are managed in a circular buffer, removing older points as new messages arrive, ensuring an up-to-date representation of the environment.


\subsection{Obstacles composition}

Solving the obstacle-avoidance constraint of (\ref{eqn:cbf_qp_with_fov})
requires to compute the values of the composite \ac{cbf} $h(\state)$, and its derivatives $\Lie{f}h(\state)$ and $\Lie{g}h(\state)$
for the current list of obstacles.
Since these are measured in the body frame, it is efficient to implement the filtering procedure in the body frame as well.
Equations (\ref{eqn:cbf_nu0}), (\ref{eqn:cbf_nu1}), and (\ref{eqn:cbf_lie_nu1}) are expressed in $\mathcal{B}$ accordingly.
The full composition procedure is described in Algorithm \ref{alg:composite_cbf}, where $\local{\pos}[1\dots n]$ denotes the array storing $\local{\pos}_1, \dots, \local{\pos}_n$.

\begin{algorithm}
\caption{Composite CBF computation}
\label{alg:composite_cbf}

\hspace*{\algorithmicindent} \textbf{Input:} $\local{\pos}[1\dots n], \local{\vel}$ \\
\hspace*{\algorithmicindent} \textbf{Output:} $\Lie{f}h$, $\Lie{g}h$, $h$ \\
\vspace*{-1em}
\begin{algorithmic}
\STATE $\Lambda \gets 0$
\STATE $\Lie{f}h \gets 0$
\STATE $\Lie{g}h \gets [0 \ 0 \ 0]$
\FOR{$i = 1$ to $n$}
    \STATE $\nu_1 \gets -2\local{\vel}^\top \local{\pos}[i] - p_0 (\|\local{\pos}[i]\|^2 - \varepsilon^2)$
    \STATE $\Lambda \gets \Lambda + \exp\left[-\kappa s\left(\nu_1/\gamma\right)\right]$
    \STATE $\lambda \gets \exp\left[-\kappa s\left(\nu_1/\gamma\right)\right] s'(\nu_1/\gamma)$
    \STATE $\Lie{f}h \gets \Lie{f}h + 2\lambda\local{\vel}^\top\left( \local{\vel} + p_0\local{\pos}[i]\right)$
    \STATE $\Lie{g}h \gets \Lie{g}h - 2\lambda\local{\pos}[i]^\top$
\ENDFOR
\STATE $h \gets -(\gamma/\kappa)\ln\Lambda $
\STATE $\Lie{f}h \gets \Lie{f}h / \Lambda$
\STATE $\Lie{g}h \gets \Lie{g}h / \Lambda$
\end{algorithmic}
\end{algorithm}

\begin{table}[t]
\centering
\caption{Summary of the range of values for each parameter and the
effect of increasing them. The behavior when the parameters are decreased
is implicitly the behavior opposed to the increasing one.}
\begin{tabular}{|c|c|l|}
\hline
\textbf{Parameter} & \multicolumn{1}{c|}{\textbf{Range}} & \multicolumn{1}{c|}{\textbf{Effect when increasing}} \\ \hline
$\varepsilon$ & $> 0$ & Larger avoidance radius \\ \hline
$\kappa$ & $[10, 100]$ & Less smooth approximation \\ \hline
$\gamma$ & $[10, 100]$ & Reacts to farther obstacles\\ \hline
$\alpha$ & $[1,3]$ & Increase filter sensitivity \\ \hline
$p_0$ & $[-3,-1]$ & Damped response \\ \hline
$\alpha_f$ & $[2,8]$ & Aggressive \ac{fov} response \\ \hline
$\tau$ & $[0.01,0.1]$ & Stronger accel smoothing \\ \hline
\end{tabular}
\label{tab:param_tuning}
\end{table}

\begin{figure}[t]
\centering
\includegraphics[width=.48\linewidth]{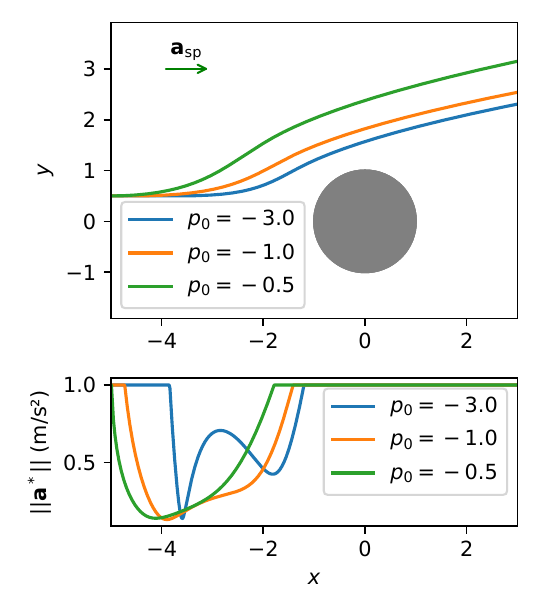}
\includegraphics[width=.48\linewidth]{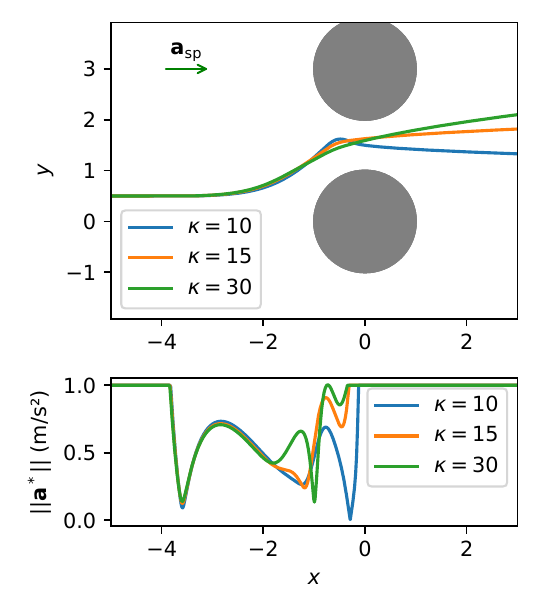}
\caption{Safety filter response for different parameter values.}
\label{fig:tuning}
\end{figure}

\begin{figure*}[t]
    \centering
    \includegraphics[width=\linewidth]{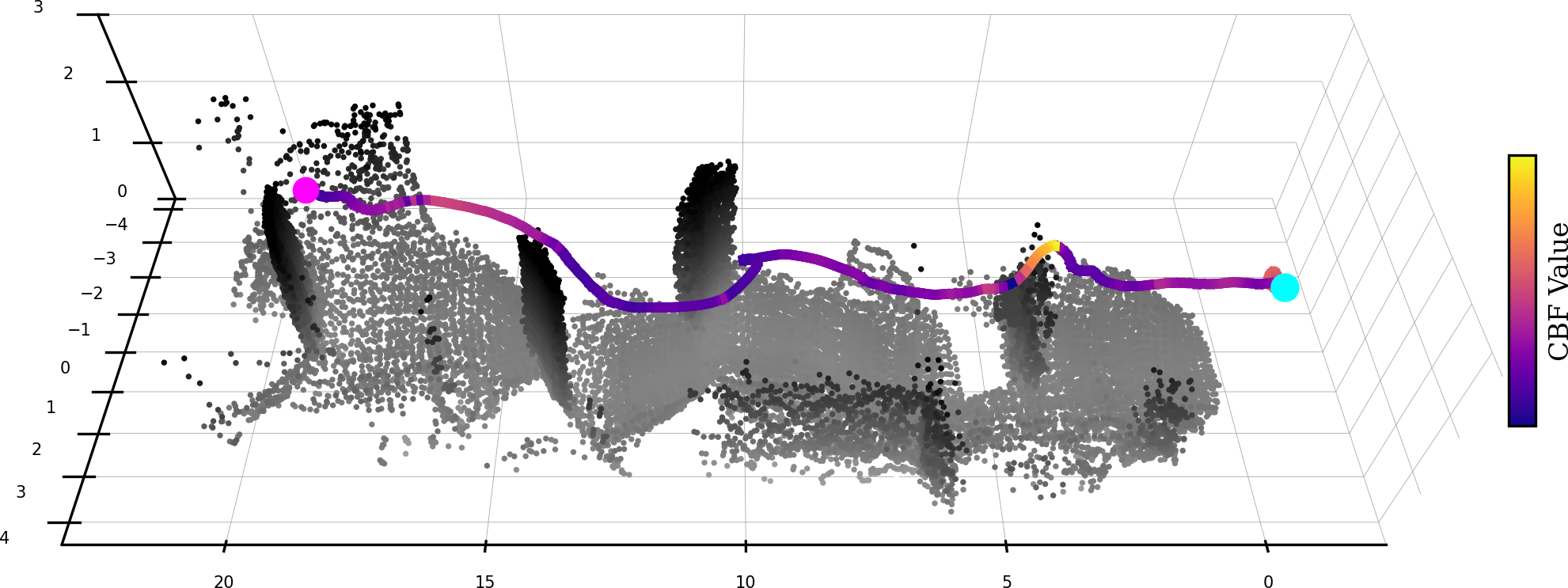}
    \caption{Visualization of the environment and flown trajectory during Experiment 1. The start and end of the mission are marked with the cyan and magenta dots, respectively. The \ac{cbf} value is generally lowest before reaching an obstacle and higher when moving in free space.}
    \label{fig:mission_adversarial}
\end{figure*}

\subsection{Safety filtering}
In the general case, with multiple \ac{fov} constraints and non-isotropic filtering of the acceleration, we utilize \texttt{qpOASES} \cite{qpoases} -- a generic online active set strategy QP solver \cite{qp_active_set} -- to solve (\ref{eqn:cbf_qp_with_fov}). The dependencies of \texttt{qpOASES} were modified to enable compilation for PX4-based micro-controllers. 

In the simplified case with no \ac{fov} constraints and isotropic filtering of the acceleration,
we directly compute the
safe acceleration setpoint $\local{\acc^*}$ from a nominal acceleration setpoint
$\local{\acc_{\textrm{sp}}}$ by solving the QP
\begin{equation*}
\begin{split}
    \local{\acc^*} = & \argmin_{\acc \in \mathbb{R}^3}
        \| \acc - \local{\acc_{\textrm{sp}}} \|^2 \\
    & \textrm{s.t.} \ \Lie{g}h(\state)\acc
        \ge - \Lie{f}h(\state) - \alpha(h(\state)).
\end{split}
\end{equation*}
This particular problem has a solution that can be computed analytically \cite{cbf_auto_vehicle} as

\begin{equation}
    \local{\acc^*} = \local{\acc_{\textrm{sp}}} + \max\{ 0, \eta(\state) \} \Lie{g}h(\state)^\top
\end{equation}
where

\small
\begin{equation}
\eta(\state)=\begin{cases}
-\frac{\Lie{f}h(\state) + \Lie{g}h(\state)\local{\acc_{\textrm{sp}}} + \alpha (h(\state))}{\|\Lie{g}h(\state)\|^2} & \mathrm{if} \
    \Lie{g}h(\state) \neq \mathbf{0}\\
0 & \mathrm{if} \ \Lie{g}h(\state) = \mathbf{0}.
\end{cases}
\end{equation}
\normalsize
In both cases, we reduce chattering (a known problem of safety filters~\cite{cbf_chattering}) by pre- and post-filtering the input $\acc_\text{sp}$ and output $\acc^*$ with the first-order low-pass filter

\begin{equation}
    \dot{\acc}_\text{filt} = \frac{1}{\tau} (\acc - \acc_\text{filt} )
\end{equation}
with the tunable parameter $\tau$. This is particularly important when the \ac{fov} constraints are present due to their intersection in the hover state, making the set of safe accelerations non-smooth.

\subsection{Parameter tuning}

Each of the parameters $\varepsilon$, $\kappa$, $\gamma$, $\alpha$, $p_0$, $\alpha_f$ and $\tau$
must be tuned according to the specific hardware setup and
application, as they directly influence the safety filter's behavior.
Table~\ref{tab:param_tuning} summarizes the range and effect of each parameter. The parameters used for the experimental study in this work are listed in Table~\ref{tab:parameters_exp}.

A brief tuning procedure of the safety filter is summarized as follows:

\begin{enumerate}
    \item Tune all low-level controllers used for acceleration tracking first, targeting a rise time of \SI{100}{\milli\second}-\SI{200}{\milli\second} for the roll and pitch axis
    \item set $\epsilon$ to the desired value, depending on the hardware.
    \item set $\gamma$ and $\kappa$ to a large value initially. Reduce kappa if more smoothing of the CBF is desired. Disable the FoV constraints.
    \item set $\alpha=2$, $p_0=-1$. Carefully reduce $p_0$ if a more aggressive response is needed. Here, $\alpha$ and $p_0$ can be interpreted as the "acceleration gain" and "velocity gain" of the filter, respectively.
    \item Enable the FoV constraints, set $\alpha_f=1$ and increase until the response is satisfactory. There should be no larger oscillations at hover.
    \item 
    Adjust the low-pass filtering gain $\tau$ to mitigate remaining high-frequency oscillations.
\end{enumerate}

An illustration of the effect of different choices of the parameters $p_0$ and $\kappa$ is given in Fig. \ref{fig:tuning}.

\begin{figure}[t]
    \centering
    \includegraphics[width=0.98\linewidth]{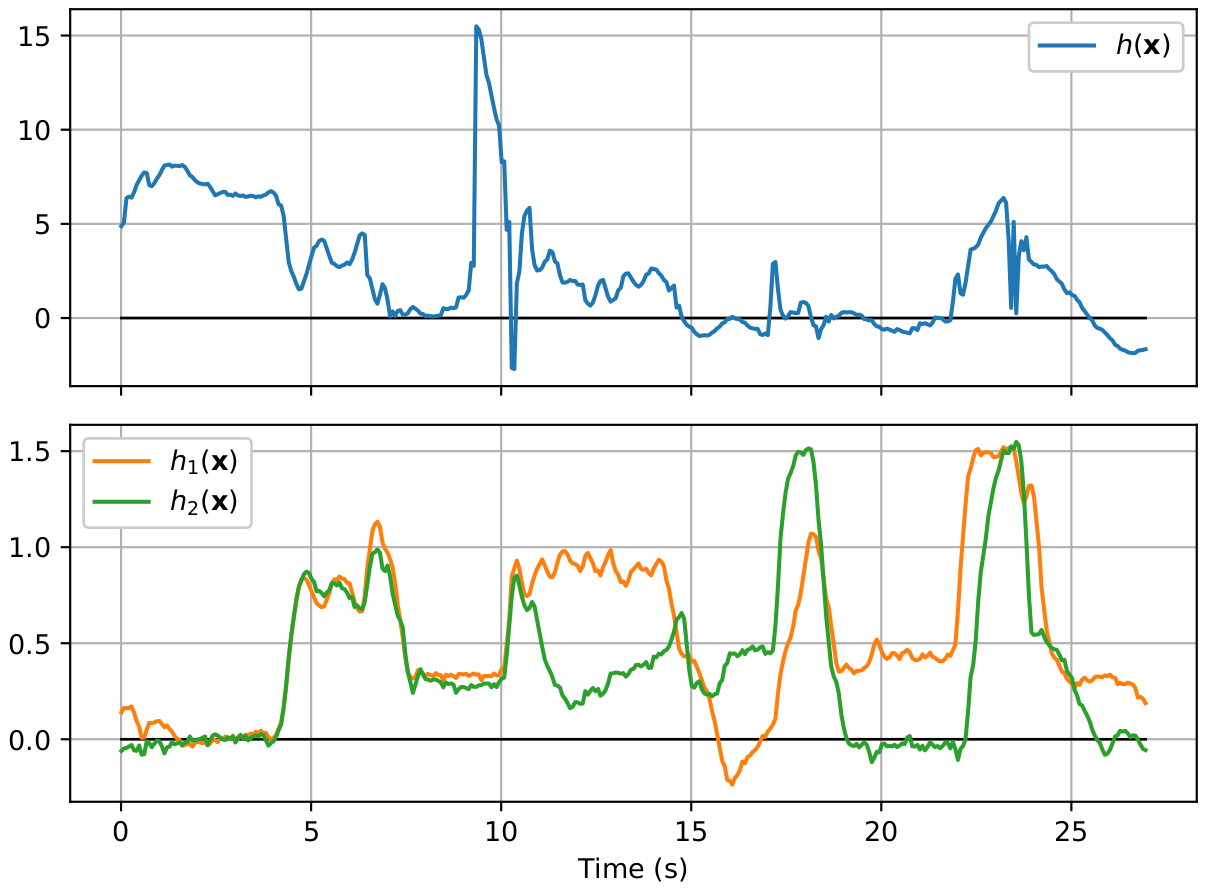}
    \caption{Values of the composite \ac{cbf} (top) and \ac{fov} \acp{cbf} (bottom) during Experiment 1.}
    \label{fig:cbf_adversarial}
\end{figure}

%% file: experiments.tex
\section{Results}\label{sec:results}

\subsection{System setup}
The safe position control architecture is evaluated with an implementation in C++ in the PX4 flight control stack. The performance of the proposed implementation is evaluated in two different use-cases. First, a user manually provides adversarial velocity commands explicitly commanding the drone to collide into obstacles. Second, an elliptic trajectory tracking is performed in the presence of obstacles.
The platform utilized to evaluate the implementation makes use of a ModalAI VOXL2 mini compute unit, which serves the role of both the onboard computer and flight controller. This enables seamless integration with entirely onboard monocular state estimation using ROVIO~\cite{rovio} and a pmd flexx 2 wide ToF depth sensor. 
The \ac{tof} range measurements are clamped at \SI{5}{m}.
A maximum of 100 obstacles are obtained from the sparsified range measurement at a rate of \SI{15}{Hz}.

On the embedded board, the total computation time for \ac{cbf} composition and solving of the analytical (unconstrained) QP scales linearly with the number of obstacles.
It takes in average \SI{50}{\mu s} with $25$ obstacles and \SI{400}{\mu s} with $200$ obstacles.
The solving of the \ac{fov}-constrained QP problem takes, on average, \SI{2}{ms} with spikes up to \SI{2.5}{ms} in iterations where one of the slack variables $\delta_{fi}$ becomes non-zero, which does not induce any significant delay in the position controller running at \SI{100}{Hz}.

The two experiments presented hereafter can be seen in the supplementary video.

\begin{table}[t]
    \centering
    \caption{Parameter values during the experiments}
    \label{tab:parameters_exp}
    \setlength{\tabcolsep}{5.5pt} 
    \renewcommand{\arraystretch}{1} 
    \begin{tabular}{|c|ccccccc|}
        \hline
        \textbf{Parameter} & $\epsilon$ & $\kappa$ & $\gamma$ & $\alpha$ & $p_0$ & $\alpha_f$ & $\tau$  \\
        \hline
        \textbf{Value} & 0.7 & 70 & 40 & 2 & -2.5 & 6 & 0.5 \\
        \hline
    \end{tabular}
    \vspace{-4ex}
\end{table}

\subsection{Experiment 1}
In the first experiment, we evaluate the safe control architecture in an adversarial tracking scenario. Here, a human operator providing velocity setpoints from a joystick tries to purposefully collide the UAV with obstacles in a university hallway. The safety filter deflects the quadrotor above and around obstacles and brings it to a full stop in front of a wall. The flown trajectory and environment are visualized in Fig. \ref{fig:mission_adversarial}. The resulting \ac{cbf} values for $h$, $h_1$ and $h_2$ are shown in Fig. \ref{fig:cbf_adversarial}. It can be seen that the \ac{cbf} are almost entirely positive, indicating the UAV remains within the safe set and avoids collisions. The values drop below the zero line only for a few instances. This can be explained by the sudden appearance of unseen obstacles, observation noise, and tracking errors present during acceleration tracking. Over the experiment, the safe controller successfully counteracts unsafe references and avoids collisions.

\subsection{Experiment 2}
\begin{figure}
    \centering
    \includegraphics[width=\linewidth]{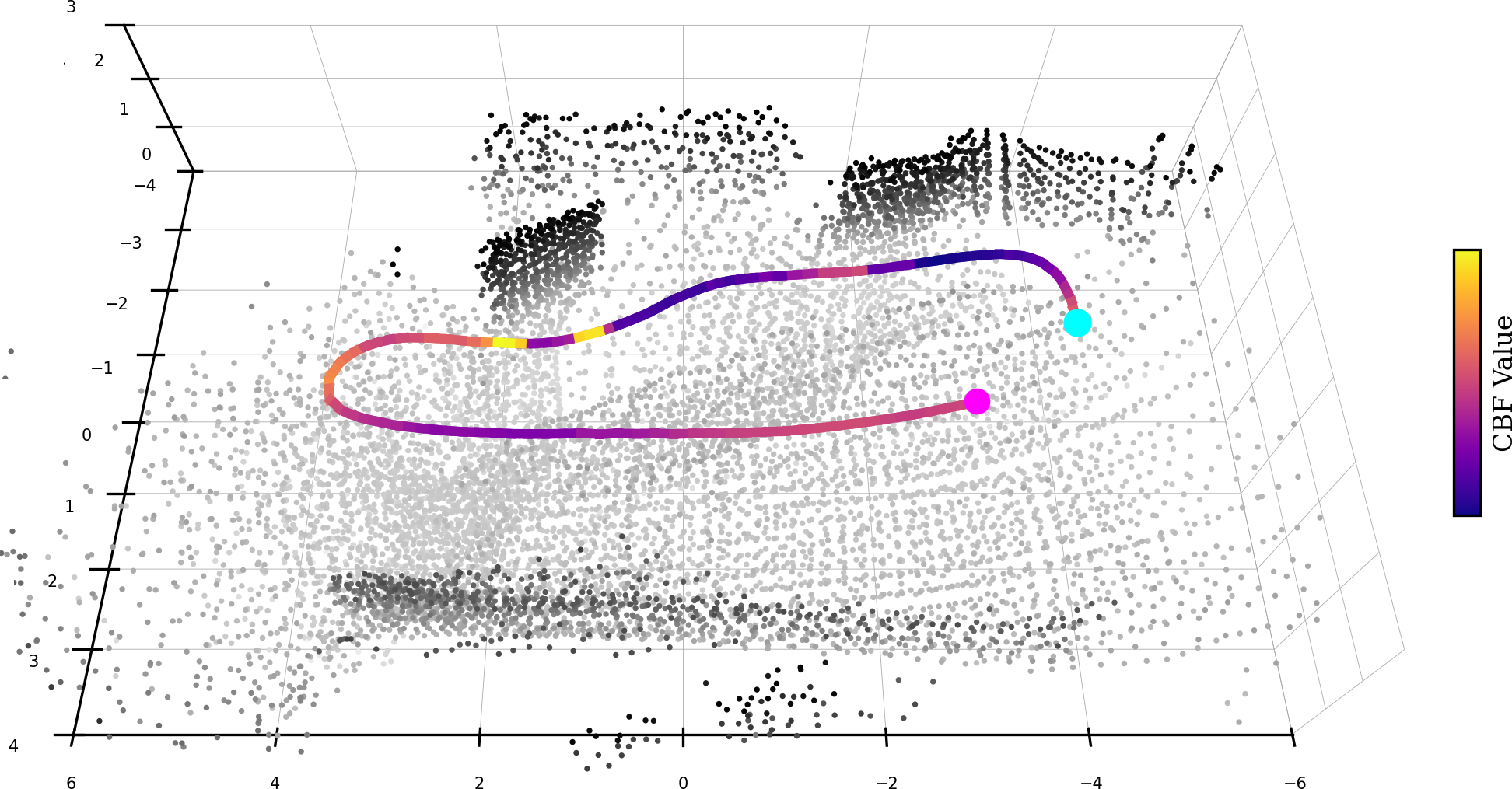}
    \caption{A short section from experiment 2. Along the path from the cyan dot to the magenta dot, the multirotor encounters a suddenly appearing obstacle. The value of $h$ is color-coded in the path. The figure shows that $h$ is smallest when approaching the two panels along the way and becomes large after losing an obstacle from view.}
    \label{fig:mission_ellipse}
\end{figure}

\begin{figure}
    \centering
    \includegraphics[width=0.98\linewidth]{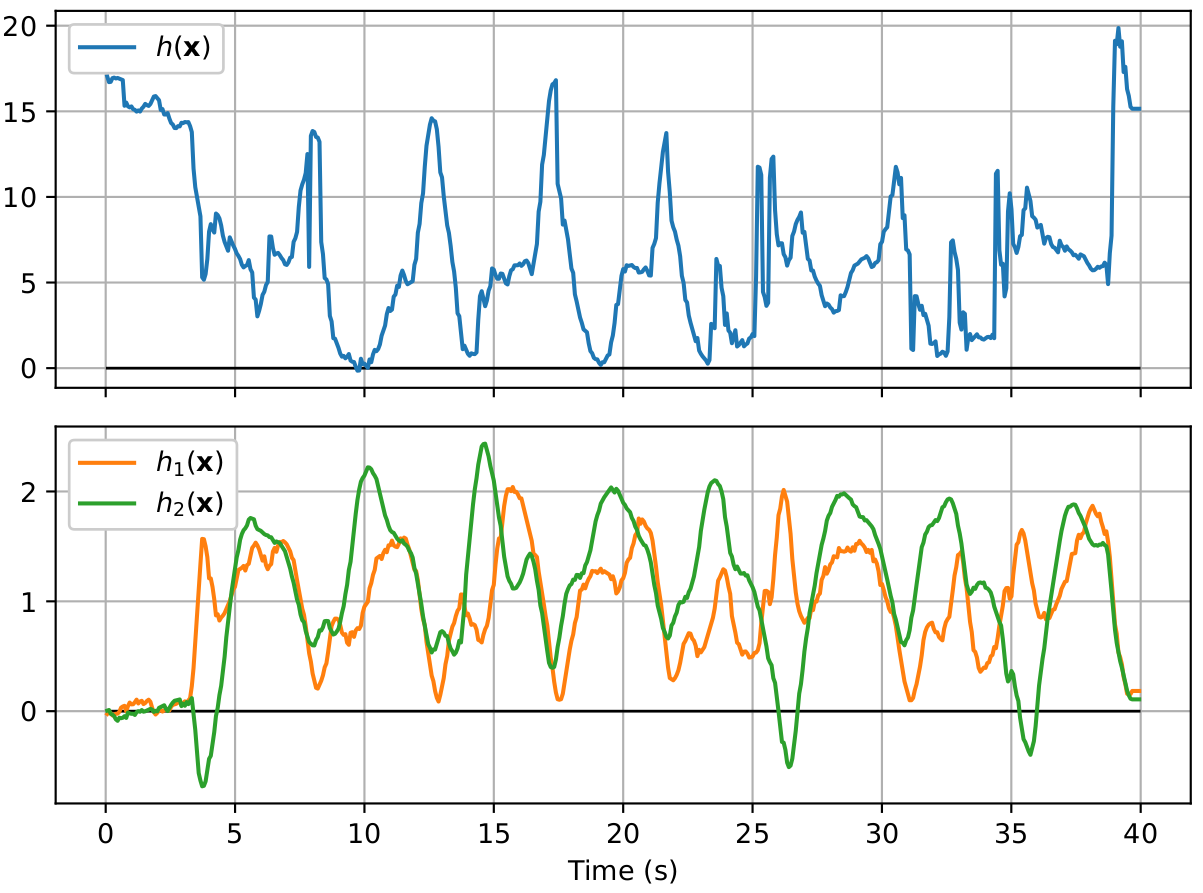}
    \caption{\ac{cbf} values during experiment 2. At around $t=20$, the panel-shaped obstacle is inserted during the third round. More spikes on the composite \ac{cbf} appear during avoidance of the new obstacle.}
    \label{fig:cbf_ellipse}
\end{figure}

In the second experiment, the robot is set to track an aggressive ellipse trajectory, achieving velocities of up to \SI{2.75}{\meter/\second} and accelerations of \SI{4}{\meter/\second^2}.
During one of the consecutive cycles, a panel is positioned on the trajectory to emulate a suddenly appearing, unmapped obstacle. The reactive safety filter alters the acceleration setpoint to avoid collision and resumes tracking the reference trajectory after the avoidance maneuver. A short excerpt of one cycle including the obstacle is shown in Fig.~\ref{fig:mission_ellipse}. The corresponding \ac{cbf} values in Fig.~\ref{fig:cbf_ellipse} show that the composite \ac{cbf} remains positive, while one of the FoV \acp{cbf} shows slight negative values repeatedly. This behavior is expected as the larger yaw rates at significant velocities at the vertex of the ellipse can cause a skidding motion, which can cause a violation of the field of view constraints. It should be noted that the additional \ac{fov} constraints deteriorate tracking performance since lateral accelerations are constrained by the corresponding \acp{cbf}. Regardless, the safe control architecture succeeds in tracking aggressive trajectories while simultaneously ensuring collision avoidance without using a consistent map.

\subsection{Limitations}
While the presented control architecture enables embedded collision avoidance in unknown environments, there are several limitations. One Limitation of the current architecture is the absence of obstacle information outside the sensor frustum, which can result in suddenly appearing or disappearing obstacles. This may cause non-smooth and jerky behavior of the safety filter in close proximity to the environment. This is a common issue of mapless methods and can be resolved using appropriate sensors with a large FoV.
One further limitation of the approach is its sensitivity to erroneous obstacle measurements. While no issues were observed in our experiments, thanks to the reliability of the ToF sensor, other sensor modalities, such as stereo depth, tend to be noisy and prone to outliers. Since the used CBF formulation is sensitive to outliers, the obstacle measurements should be filtered appropriately.

%% file: conclusion.tex
\section{Conclusion}\label{sec:conclusions}

This work proposes a safe control architecture for multirotors based on CBFs, which enforces collision-free navigation in cluttered environments. The architecture leverages local observations only and serves as a last-resort safety layer, designed to seamlessly integrate with any task-oriented method.
No mapping is assumed and only directly visible obstacles 
detected from a low-resolution range sensor are used in a safety filter
leveraging a scalable composite CBF.
The embedded implementation of the safety filter is open-sourced in a fork of the PX4 autopilot framework as part of the default position and velocity controller module, facilitating its use in academic or industrial fields.
We demonstrated the effectiveness and usability of the approach in two hardware experiments in which collision avoidance is achieved during a trajectory tracking task, as well as against adversarial velocity commands.

Future work should investigate creating short-term memory at constraint level to alleviate the non-smooth behavior due to disappearing obstacles, and investigate passivity-based approaches for safety to dampen the response in extremely cluttered environments.

%% file: main.bbl
\begin{thebibliography}{10}
\providecommand{\url}[1]{#1}
\csname url@samestyle\endcsname
\providecommand{\newblock}{\relax}
\providecommand{\bibinfo}[2]{#2}
\providecommand{\BIBentrySTDinterwordspacing}{\spaceskip=0pt\relax}
\providecommand{\BIBentryALTinterwordstretchfactor}{4}
\providecommand{\BIBentryALTinterwordspacing}{\spaceskip=\fontdimen2\font plus
\BIBentryALTinterwordstretchfactor\fontdimen3\font minus \fontdimen4\font\relax}
\providecommand{\BIBforeignlanguage}[2]{{%
\expandafter\ifx\csname l@#1\endcsname\relax
\typeout{** WARNING: IEEEtran.bst: No hyphenation pattern has been}%
\typeout{** loaded for the language `#1'. Using the pattern for}%
\typeout{** the default language instead.}%
\else
\language=\csname l@#1\endcsname
\fi
#2}}
\providecommand{\BIBdecl}{\relax}
\BIBdecl

\bibitem{acevedo2013cooperative}
J.~J. Acevedo, B.~C. Arrue, I.~Maza, and A.~Ollero, ``Cooperative large area surveillance with a team of aerial mobile robots for long endurance missions,'' \emph{Journal of Intelligent \& Robotic Systems}, vol.~70, pp. 329--345, 2013.

\bibitem{popovic2020informative}
M.~Popovi{\'c}, T.~Vidal-Calleja, G.~Hitz, J.~J. Chung, I.~Sa, R.~Siegwart, and J.~Nieto, ``An informative path planning framework for uav-based terrain monitoring,'' \emph{Autonomous Robots}, vol.~44, no.~6, pp. 889--911, 2020.

\bibitem{dang2019graph}
T.~Dang, F.~Mascarich, S.~Khattak, C.~Papachristos, and K.~Alexis, ``Graph-based path planning for autonomous robotic exploration in subterranean environments,'' in \emph{2019 IEEE/RSJ International Conference on Intelligent Robots and Systems (IROS)}.\hskip 1em plus 0.5em minus 0.4em\relax IEEE, 2019, pp. 3105--3112.

\bibitem{zhu2021online}
H.~Zhu, J.~J. Chung, N.~R. Lawrance, R.~Siegwart, and J.~Alonso-Mora, ``Online informative path planning for active information gathering of a 3d surface,'' in \emph{2021 IEEE International Conference on Robotics and Automation (ICRA)}.\hskip 1em plus 0.5em minus 0.4em\relax IEEE, 2021, pp. 1488--1494.

\bibitem{Sucan12}
I.~A. Sucan, M.~Moll, and L.~E. Kavraki, ``The open motion planning library,'' vol.~19, no.~4, pp. 72--82, 2012.

\bibitem{Tranzatto22}
M.~Tranzatto, T.~Miki, M.~Dharmadhikari, L.~Bernreiter, M.~Kulkarni, F.~Mascarich, O.~Andersson, S.~Khattak, M.~Hutter, R.~Siegwart, and K.~Alexis, ``Cerberus in the darpa subterranean challenge,'' vol.~7, no.~66, 2022.

\bibitem{Ebadi23}
K.~Ebadi, L.~Bernreiter, H.~Biggie, G.~Catt, Y.~Chang, A.~Chatterjee, C.~E. Denniston, S.-P. Deschênes, K.~Harlow, S.~Khattak, L.~Nogueira, M.~Palieri, P.~Petráček, M.~Petrlík, A.~Reinke, V.~Krátký, S.~Zhao, A.-a. Agha-mohammadi, K.~Alexis, C.~Heckman, K.~Khosoussi, N.~Kottege, B.~Morrell, M.~Hutter, F.~Pauling, F.~Pomerleau, M.~Saska, S.~Scherer, R.~Siegwart, J.~L. Williams, and L.~Carlone, ``Present and future of {SLAM} in extreme environments: The {DARPA} subt challenge,'' \emph{Transactions on Robotics}, vol.~40, pp. 936--959, 2024.

\bibitem{Florence18}
P.~Florence, J.~Carter, J.~Ware, and R.~Tedrake, ``Nanomap: Fast, uncertainty-aware proximity queries with lazy search over local 3d data,'' in \emph{2018 IEEE international conference on robotics and automation (ICRA)}, 2018, pp. 7631--7638.

\bibitem{Gao19}
F.~Gao, W.~Wu, W.~Gao, and S.~Shen, ``Flying on point clouds: Online trajectory generation and autonomous navigation for quadrotors in cluttered environments,'' \emph{JOURNAL of Field Robotics}, vol.~36, no.~4, pp. 710--733, 2019.

\bibitem{Zhou22}
X.~Zhou, X.~Wen, Z.~Wang, Y.~Gao, H.~Li, Q.~Wang, T.~Yang, H.~Lu, Y.~Cao, C.~Xu, and F.~Gao, ``Swarm of micro flying robots in the wild,'' vol.~7, no.~66, 2022.

\bibitem{cbf_qp}
A.~D. Ames, X.~Xu, J.~W. Grizzle, and P.~Tabuada, ``Control barrier function based quadratic programs for safety critical systems,'' \emph{IEEE Transactions on Automatic Control}, vol.~62, no.~8, pp. 3861--3876, 2017.

\bibitem{Wabersich21}
K.~P. Wabersich and M.~N. Zeilinger, ``A predictive safety filter for learning-based control of constrained nonlinear dynamical systems,'' \emph{Automatica}, vol. 129, p. 109597, 2021.

\bibitem{3d_quadrotor_safe_control}
G.~Wu and K.~Sreenath, ``Safety-critical control of a 3d quadrotor with range-limited sensing,'' 10 2016, p. V001T05A006.

\bibitem{mc_cbf_backstepping}
J.~Kim and Y.~Kim, ``Safe control synthesis for multicopter via control barrier function backstepping,'' in \emph{2023 62nd IEEE Conference on Decision and Control (CDC)}, 2023, pp. 8720--8725.

\bibitem{cbf_cone}
M.~Tayal, R.~Singh, J.~Keshavan, and S.~Kolathaya, ``Control barrier functions in dynamic uavs for kinematic obstacle avoidance: A collision cone approach,'' in \emph{2024 American Control Conference (ACC)}, 2024, pp. 3722--3727.

\bibitem{softmax_cbf}
\BIBentryALTinterwordspacing
A.~Safari and J.~B. Hoagg, ``Time-varying soft-maximum barrier functions for safety in unmapped and dynamic environments,'' 2024. [Online]. Available: \url{https://arxiv.org/abs/2409.01458}
\BIBentrySTDinterwordspacing

\bibitem{learning_cbf}
C.~Dawson, B.~Lowenkamp, D.~Goff, and C.~Fan, ``Learning safe, generalizable perception-based hybrid control with certificates,'' \emph{IEEE Robotics and Automation Letters}, vol.~7, no.~2, pp. 1904--1911, 2022.

\bibitem{neural_cbf}
\BIBentryALTinterwordspacing
M.~Harms, M.~Kulkarni, N.~Khedekar, M.~Jacquet, and K.~Alexis, ``Neural control barrier functions for safe navigation,'' \emph{2024 IEEE/RSJ International Conference on Intelligent Robots and Systems (IROS)}, pp. 10\,415--10\,422, 2024. [Online]. Available: \url{https://api.semanticscholar.org/CorpusID:271533429}
\BIBentrySTDinterwordspacing

\bibitem{dyn_cbf_mpc}
Z.~Jian, Z.~Yan, X.~Lei, Z.~Lu, B.~Lan, X.~Wang, and B.~Liang, ``Dynamic control barrier function-based model predictive control to safety-critical obstacle-avoidance of mobile robot,'' in \emph{2023 IEEE International Conference on Robotics and Automation (ICRA)}, 2023, pp. 3679--3685.

\bibitem{composite_cbf_nav}
\BIBentryALTinterwordspacing
M.~Harms, M.~Jacquet, and K.~Alexis, ``Safe quadrotor navigation using composite control barrier functions,'' \emph{2025 IEEE International Conference on Robotics and Automation (ICRA)}, 2025. [Online]. Available: \url{https://arxiv.org/abs/2502.04101}
\BIBentrySTDinterwordspacing

\bibitem{cbf_theory_app}
A.~D. Ames, S.~Coogan, M.~Egerstedt, G.~Notomista, K.~Sreenath, and P.~Tabuada, ``Control barrier functions: Theory and applications,'' in \emph{2019 18th European Control Conference (ECC)}, 2019, pp. 3420--3431.

\bibitem{exp_cbf}
Q.~Nguyen and K.~Sreenath, ``Exponential control barrier functions for enforcing high relative-degree safety-critical constraints,'' in \emph{2016 American Control Conference (ACC)}, 2016, pp. 322--328.

\bibitem{composing_cbfs}
T.~G. Molnar and A.~D. Ames, ``Composing control barrier functions for complex safety specifications,'' \emph{IEEE Control Systems Letters}, vol.~7, pp. 3615--3620, 2023.

\bibitem{meier2015px4}
L.~Meier, D.~Honegger, and M.~Pollefeys, ``Px4: A node-based multithreaded open source robotics framework for deeply embedded platforms,'' in \emph{2015 IEEE international conference on robotics and automation (ICRA)}.\hskip 1em plus 0.5em minus 0.4em\relax IEEE, 2015, pp. 6235--6240.

\bibitem{mahony2012multirotor}
R.~Mahony, V.~Kumar, and P.~Corke, ``Multirotor aerial vehicles: Modeling, estimation, and control of quadrotor,'' \emph{IEEE robotics \& automation magazine}, vol.~19, no.~3, pp. 20--32, 2012.

\bibitem{qpoases}
H.~Ferreau, C.~Kirches, A.~Potschka, H.~Bock, and M.~Diehl, ``{qpOASES}: A parametric active-set algorithm for quadratic programming,'' \emph{Mathematical Programming Computation}, vol.~6, no.~4, pp. 327--363, 2014.

\bibitem{qp_active_set}
H.~Ferreau, H.~Bock, and M.~Diehl, ``An online active set strategy to overcome the limitations of explicit mpc,'' \emph{International Journal of Robust and Nonlinear Control}, vol.~18, no.~8, pp. 816--830, 2008.

\bibitem{cbf_auto_vehicle}
A.~Alan, A.~J. Taylor, C.~R. He, A.~D. Ames, and G.~Orosz, ``Control barrier functions and input-to-state safety with application to automated vehicles,'' \emph{IEEE Transactions on Control Systems Technology}, vol.~31, no.~6, pp. 2744--2759, 2023.

\bibitem{cbf_chattering}
F.~P. Bejarano, L.~Brunke, and A.~P. Schoellig, ``Multi-step model predictive safety filters: Reducing chattering by increasing the prediction horizon,'' in \emph{2023 62nd IEEE Conference on Decision and Control (CDC)}, 2023, pp. 4723--4730.

\bibitem{rovio}
M.~Bloesch, S.~Omari, M.~Hutter, and R.~Siegwart, ``Robust visual inertial odometry using a direct ekf-based approach,'' in \emph{2015 IEEE/RSJ International Conference on Intelligent Robots and Systems (IROS)}, 2015, pp. 298--304.

\end{thebibliography}
